\documentclass[10pt,twocolumn,letterpaper]{article}

\usepackage{iccv}
\usepackage{times}
\usepackage{epsfig}
\usepackage{graphicx}
\usepackage{amsmath}
\usepackage{amssymb}

\usepackage[pagebackref=true,breaklinks=true,colorlinks,bookmarks=false]{hyperref}

\iccvfinalcopy 

\def\iccvPaperID{4339} 

\ificcvfinal\fi
\usepackage[utf8]{inputenc}

\usepackage{adjustbox}
\usepackage{xcolor}
\definecolor{newnew}{rgb}{0.1, 0.6, 0.3}
\definecolor{new}{rgb}{0, 0, 0}
\definecolor{AGred}{rgb}{1, 0, 0}
\definecolor{inprogress}{rgb}{0, 0, 0}
\usepackage{makecell}
\usepackage{booktabs}       
\newcommand{\squeezeup}{\vspace{-2mm}}

\begin{document}

\title{ALADIN: All Layer Adaptive Instance Normalization \\for Fine-grained Style Similarity}

\author{Dan Ruta\textsuperscript{1} \and Saeid Motiian\textsuperscript{2} \and Baldo Faieta\textsuperscript{2} \and Zhe Lin\textsuperscript{2} \and Hailin Jin\textsuperscript{2} \and Alex Filipkowski\textsuperscript{2} \and Andrew Gilbert\textsuperscript{1} \and John Collomosse\textsuperscript{1,2} \and
$^1$CVSSP, University of Surrey\\

\and
$^2$Adobe Research\\

}

\twocolumn[{%
\renewcommand\twocolumn[1][]{#1}%
\maketitle
\vspace{-20pt}

\begin{center}
    \centering
    \includegraphics[width=0.8\linewidth]{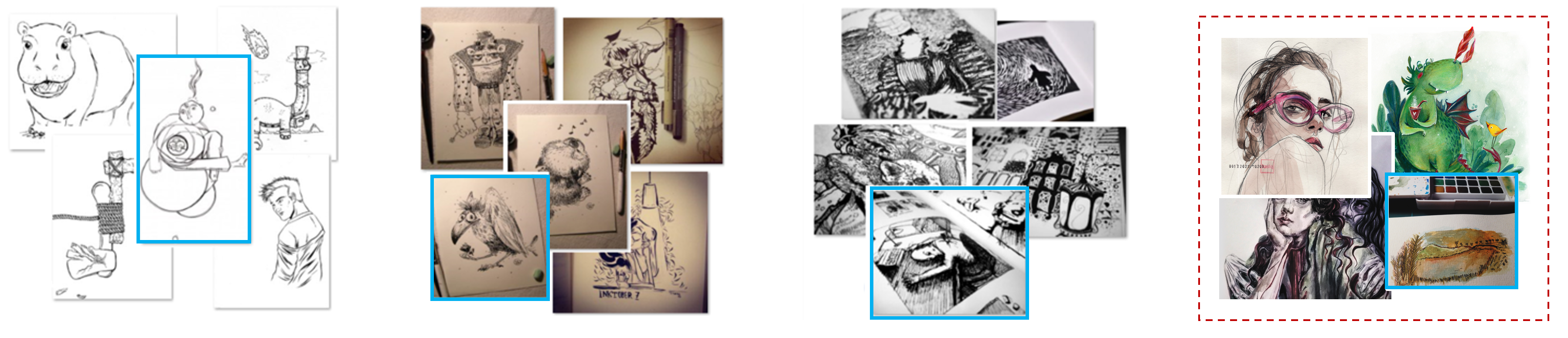}
\end{center}%
    Figure 1. ALADIN enables style-based visual search, by learning a fine-grained embedding for artistic style similarity.  Left: Three queries (blue) and their results \ie nearest neighbours in the ALADIN search embedding.  Each result set exhibits fine-grained style coherence; a consistent style of sketching.  Right (red box): a search in a prior style search embedding \cite{Collomosse2017} returns results with only coarse-grained style coherence; all are different variants of watercolor style.
\vspace{20pt}
}
]
\setcounter{figure}{1}

\ificcvfinal\thispagestyle{empty}\fi

\begin{abstract}

      We present ALADIN (All Layer AdaIN); a novel architecture for searching images based on the similarity of their artistic style.  Representation learning is critical to visual search, where distance in the learned search embedding reflects image similarity.  Learning an embedding that discriminates fine-grained variations in style is hard, due to the difficulty of defining and labelling style.  ALADIN takes a weakly supervised approach to learning a representation for fine-grained style similarity of digital artworks, leveraging BAM-FG, a novel large-scale dataset of user generated content groupings gathered from the web. ALADIN sets a new state of the art accuracy for style-based visual search over both coarse labelled style data (BAM)  and BAM-FG; a new 2.62 million image dataset of 310,000 fine-grained style groupings also contributed by this work.
\end{abstract}

\section{Introduction}
Digital artwork spans a broad range of content depicted in diverse visual styles.  Learning a representation suitable for searching artwork based on visual style is an open challenge, particularly when discriminating between subtle, {\em fine-grained} \cite{finegrained_cvpr11, finegrained_ieee17, finegrained_cvpr20} variations in style. This is due to the difficulties of both (i) defining a suitable fine-grained ontology to label styles and (ii) the expert annotation task. Research to date has therefore focused upon coarse-grain discrimination of a limited number of styles \cite{Karayev2014,Collomosse2017}.

Artistic style is the distinctive appearance of an artwork; \ie how an artist has depicted their subject matter \cite{styledef}.  Style may be characterized, non-exhaustively, by visual attributes such as the texture, strokes, media, shading, or layout of an artwork; the challenge of identifying a complete list of attributes is long-standing and unsolved \cite{styleont}. Our core contribution is to learn {\em fine-grained} artistic style similarity (Fig. 1) and do so via a weakly supervised approach that does not rely upon explicit labelling of style or style attributes in images.  Our technical contributions are three-fold:

{\bf 1. ALADIN Fine-grained Style Embedding.}  We propose ALADIN; a novel architecture to learn a search embedding for image style.  ALADIN is an encoder-decoder (E-D) network, that pools Adaptive Instance Normalization (AdaIN) statistics across its style encoder layers to learn a discriminative search embedding capable of both discriminating subtle fine-grained of style (\eg variations in sketching style) as well as coarse-grained styles (\eg sketch, watercolor, etc.).  Image stylization networks \cite{HuangAdaIn2017,google_adain,munit} have previously used AdaIN for style transfer, but these perform poorly for measuring similarlity (c.f. subsec~\ref{sec:coarsefinegrained}).  For the first time, ALADIN explicitly disentangles content and style within an E-D network to show how AdaIN may be harnessed for fine-grained style search.

{\bf 2. Behance Artistic Media Fine-Grained (BAM-FG) dataset. }
We contribute a new 2.62 million image dataset of artwork within 310K fine-grained style groupings, gathered from a creative portfolio website (Behance.net).  Digital artists publish to Behance.net in micro-collections (`projects') comprising images of a related visual theme.  On the assumption that image co-occurrence within these groups implies a weak cue for style similarity, we sample millions of such co-occurrences. Further, we partition and clean this noisy co-occurrence data via a large-scale crowd annotation task in which distinct style-coherent sub-groups of images within projects are identified with high consensus (yielding 1.62 million images and 135K groups).


{\bf 3. Weakly supervised learning of fine-grained style.}  We present the first study into representation learning for {\em fine-grained} artistic style similarity, taking a weakly supervised approach.  Prior style-based visual search learns only {\em coarse-grain style} discrimination directly from {\em explicitly labelled} data (\eg via proxy classification tasks \cite{Karayev2014} or  deep metric learning  \cite{Collomosse2017}).   We train ALADIN using supervised contrastive learning  \cite{simclr,khosla2020supervised} to achieve a state of the art performance at both coarse and fine-grained style search without requiring any explicit coarse of fine-grained categorization of image style.


Note that we distinguish between noisy and weak supervision.  The supervision is weak because there is no fine-grained style ontology to label images explicitly; instead, a weak proxy via implicit project groupings is the basis for learning.  These groupings may be noisy or be cleaned up via crowd annotation, but supervision remains weak, as the data is not explicitly labelled to fine-grained styles.  We show via objective and subjective user trials that raw project groupings are sufficient to train a state of the art model for fine-grained style similarity.  Our cleaned data is used both for evaluating and enhancing fine-grained style discrimination of ALADIN.

\begin{figure*}
\begin{center}

\includegraphics[width=0.85\linewidth,height=5.5cm]{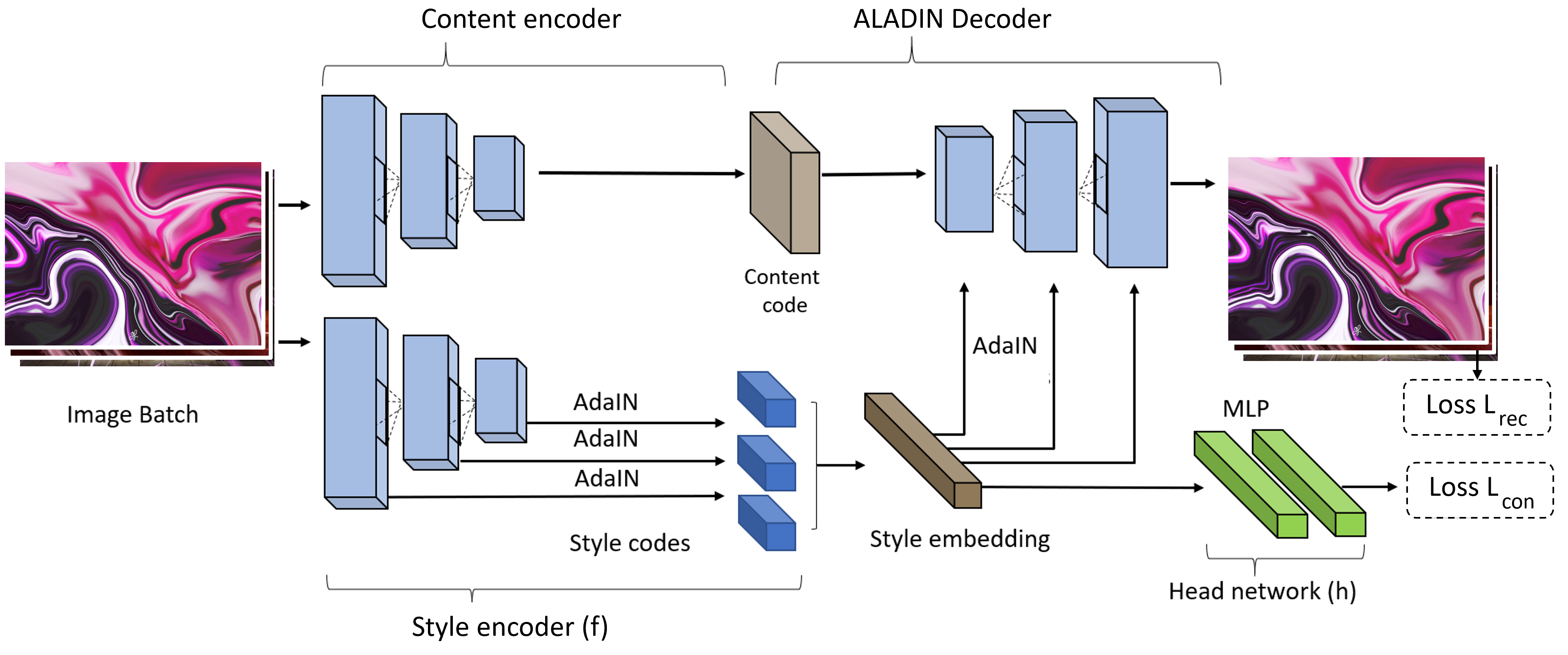}
\end{center}
   \caption{Proposed ALADIN architecture for learning a fine-grained style embedding.  ALADIN uses a multiple stage encoder where AdaIN values are aggregated from each encoder layer and passed to the corresponding decoder stages.  A concatenation of AdaIN features from encoder layers on the style branch is trained via a dual reconstruction ($L_\mathrm{rec}$) and constrastive loss ($L_\mathrm{con}$)  under weak supervision from project group co-membership.  The style encode/decoder backbone may take the form of several convolutional layers (ALADIN-S) or VGG-16 backbone (ALADIN-L).
   }
\label{fig:unet}
\end{figure*}


\section{Related Work}

Visual style has been researched primarily from the perspectives of synthesis (\eg style transfer) and classification.  Early style transfer work learned visual analogies from photo-artwork pairs \cite{Hertzmann2001,Lee2000}.  More recently, deep representations enabled stylization from unpaired data \cite{gatys}. Notably, Gatys \etal computed the Grammian across layers of a pre-trained model (\eg VGG-19 \cite{Simonyan2015}) to abstract content from style in diverse imagery; a representation that has been exploited for texture description \cite{Lin2016}, style-coherent in-painting \cite{Gilbert2018}, and fast neural style transfer (NST) via style-specific encoder-decoder networks \cite{Ulyanov2016,Johnson2016}.  Extensions to multi-scale \cite{Wang2017} and video \cite{Ruder2016} NST were later presented. Variants of this representation (\eg cosine-of-Grammian) have been explored to represent style \cite{gmdotcos}, and the concepts of image analogy and NST combined for style transfer~\cite{liao2017visual,upchurch2016z}.   Instance normalization was proposed to enhance the quality of style transfer \cite{Ulyanov2017}, and building upon this, mean-variance statistics (AdaIN) between content and style features \cite{HuangAdaIn2017,google_adain}.
Recently unsupervised style transfer was enabled via MUNIT \cite{munit}, which disentangled content and style via AdaIN \cite{HuangAdaIn2017}; learning a latent code for style without labelled data within an encoder-decoder (E-D) architecture.  A similar architecture was later used to swap style between images \cite{Park2020}. These approaches can be considered to embed the notion of style within architectural design choices.  We explore the complementary problem of learned search representations via E-D models.

By contrast, trained representations for style learn disentangled embeddings for content and style explicitly via deep metric learning, supervised using triplets  \cite{Collomosse2017}.  Their work leverages an extensive public dataset of coarsely labelled digital artwork (Behance Artistic Media -- BAM \cite{bam}). Smaller collections of labelled data have been used to supervise the classification of style \cite{Zujovic2009,Karayev2014}, or product designs \cite{Bell2015} and even painters \cite{Cetinic2013} and artistic genres \cite{Shamir2010}. All these approaches use direct supervision of coarse-grain class labels on style, \eg of fine-art \cite{omniart}.  A related field models image aesthetics using votes on social media \cite{Marchesotti2013}, and recently the link between style and emotion is explored \cite{artemis,wikiartemo}.  Generative adversarial networks (GAN) such as cycle-consistent GAN \cite{cyclegan} have been trained to map images from one domain to another, including between styles, and require labelled sets of (unpaired) images.  Recently StyleGAN \cite{styleGAN} explored the injection of a learned style-code at multiple stages in a convolutional encoder.
Our work tackles style representation learning with weak labels without requiring explicit class supervision while focusing (being able to) discriminate fine-grained styles (Fig. 1).

\section{Learning Fine-Grained Style Similarity}

Our goal is to learn a fine-grained representation of style from a weak proxy (group co-membership) rather than direct supervision under labels from a style ontology.

\subsection{ALADIN Architecture}
\label{sec:arch}

We propose ALADIN; All-Layer ADaptive Instance Normalization designed for content and style disentanglement using adaptive instance normalization (AdaIN eq. \ref{eq:adain}) \cite{HuangAdaIn2017,google_adain} in an encoder-decoder (E-D) network (Fig. \ref{fig:unet}). AdaIN has been applied to neural style transfer (NST), where content is disentangled from style to enable modification of the style code \cite{munit,Park2020} before recombination of the two.  However, style codes for these models perform very poorly when used as a search embedding (c.f. subsec.~\ref{sec:coarsefinegrained}).  ALADIN implements a disentangled E-D network design but extracts the mean and variation (as per AdaIN) values of activations from several style encoder layers in the latent code (`style embedding').  By training both for high reconstruction fidelity and with a  contrastive approach to encourage metric properties in this latent space, we show ALADIN can learn  a {\em search embedding} suitable for fine-grained style search. 


ALADIN comprises dual encoder branches: (i) content encoder; (ii) style encoder. The content encoder uses 4 convolutional layers to downsample image features into a series of semantically-focused feature maps; instance normalization is applied to each layer. The  style encoder branch comprises three convolutional layers, comprising of 64, 128, 256 filters, respectively. We extract the style information using the AdaIN mean and variance statistics of the feature maps, rather than a fully connected (FC) layer, as this encoder is used only for style extraction. The style code is composed of twice as many values as filters in the encoder, with a mean and variance of each output feature map.  Fig. \ref{fig:unet} shows the ALADIN backbone built to include adaptive instance normalization (AdaIN) on the style encoder stages and the aggregation of activations (`style codes') from multiple encoder stages via concatenation to produce an 896-D search embedding.

The decoder mirrors the encoder shape, such that the style code can be split back into the same sized segments and applied to decoder filters. Both this mirroring and the multi-layer encoding of AdaIN differ from stylization E-D networks (\eg MUNIT \cite{munit}) and enable a more effective search embedding to be learned from the resulting style codes. In the decoder stylization stage, pairs of [mean $\mu(.)$ , variance $\sigma^2(.)$] values from style encoder layer activations (\eg x) are applied to activations from a mirrored layer in the ALADIN decoder (\eg y), following eq. \ref{eq:adain}.

\squeezeup
\squeezeup
\begin{eqnarray}
\label{eq:adain}
AdaIN(x,y) = \sigma^2(y) \left( \frac{x-\mu(x)}{\sigma^2(x)} \right) + \mu(y)
\end{eqnarray}

For both ALADIN-S and ALADIN-L variants we apply a multi-layer perceptron (MLP) with a hidden layer of size 512 and $L_2$ normalized output vector 128-D to the encoded embedding feature $f(.)$.  We refer to this as projection network as $h(f(.))$ and later explore the efficacy of computing the loss on this projection embedding rather than the learned style embedding (c.f. Tbl.~\ref{tab:training_types}).

\subsection{Training with Implicit Project Groups}
\label{sec:train}

We train ALADIN using a variant of supervised contrastive learning adapted with logit accumulation (sec. \ref{sec:logit_accum}) to enable larger batch sizes. Training is weakly supervised, eschewing explicit style labels for the similarity in style implied by the co-presence of images in projects (groups) sampled from Behance.net creative portfolios (c.f. subsec ~\ref{sec:dataset}).

Classic approaches to deep metric learning for search exploit pair-wise comparison (\eg via triplet loss) to encourage correlation between visual similarity and embedding proximity.  For a given training set $T$, a project group $G \subset T$ is selected at random, and an image $a$ picked at random from within $G$ as the `anchor'. The remaining images within that group form positive examples $G^+ = G \setminus \{a\}$, and an equal number of negative samples are selected from other project groups $G^- \subset T \setminus G^+$.  For a given minibatch $B$ triplets $(a,p,n)$ are formed where $p \in G^+$ and $n \in G^-$, and backpropagation applied to minimize   $\sum_{(a,p,n) \in B} \left[ \epsilon + |f(a)-f(p)|_2 - |f(n)-f(p)|_2 \right]_+$, where $|.|_2$ denotes the $L_2$ loss and $\epsilon$ a small margin.

Recently, contrastive learning has shown improved performance using larger batch sizes.  We form a minibatch $B$ by sampling pairs of images $\{a,p\} \in G_i$ for $i=[1,N]$ groups (we use $N=1024$ groups). Thus a batch comprises $2N$ images, $B=\{b_1,b_2,...,b_{2N}\}$, where  $b_{2i}$ and $b_{2i-1}$  come from the same group. For a given image $b_i$ we therefore have a positive group $G^+_i=\{b_p\}$ and a negative group $G^-_i=B \setminus \{b_i ~ b_p\}$ for which the loss is:
\squeezeup
\begin{eqnarray}
\begin{split}
L_{\mathrm{con}}(B)=  \sum_{i=1}^{2N} L(i)  \mathrm{ ~~, where} \\
L(i) =  - \log \sum_{p \in G^+_i}  \frac{ \exp(f(i) \cdot f(p) / \tau)}{ \sum_{n \in G^-_i} \exp\left(f(i) \cdot f(n) / \tau\right)}
\end{split}
\end{eqnarray}

where $\tau>0$ is a temperature parameter as in the self-supervised SimCLR \cite{simclr}.

ALADIN is an encoder-decoder network, so we employ a dual loss comprising also a reconstruction loss term $L_{\mathrm{rec}}$:
\squeezeup
\begin{eqnarray}
L_{\mathrm{rec}}(B)&=& \sum_{b \in B} \left| f(b)-b \right|. \label{eq:recon}\\
L_{\mathrm{total}}(B)&=& L_{\mathrm{con}}(B) + \lambda L_{\mathrm{rec}}(B), \label{eq:dual}
\end{eqnarray}
where $|.|$ denotes the $L_1$ loss and we weight the reconstruction loss $\lambda=10^{-2}$.

\subsubsection{Logit Accumulation}
\label{sec:logit_accum}

Training ALADIN with large batches $|B|=1024$ is impractical on contemporary GPUs (without resorting to federated or distributed compute) due to the extensive VRAM requirements. To combat this issue, we propose a \textit{logit accumulation} strategy.  The large batch size is first split into several smaller chunks (\textit{gradient} batch size) for which the model in inference mode generates logits. Once the \textit{target} batch size of 1024 is reached, these are concatenated and used for computing the contrastive loss. Backpropagation is carried out to compute gradients, stopping and retaining them at the logits level. The original chunks are next passed through the model again one by one, storing gradients in the model. The logit gradients corresponding to samples in a given chunk are backpropagated through the model before finally applying the gradients to the weights. When all chunks have been re-forwarded through the model the batch ends (see sup-mat. for visualizations of this process). With this technique, a single GPU with 12GB VRAM can fit the required batch size of 1024. Multiple GPUs could, in theory, handle larger batches as the bottle-beck becomes storing the MLP head gradients in VRAM, which may be parallelized across GPUs.

\subsection{BAM-FG: Behance Artistic Media Fine-Grained dataset}
\label{sec:dataset}
\begin{figure}[t!]
    \centering
    \includegraphics[width=1.0\linewidth]{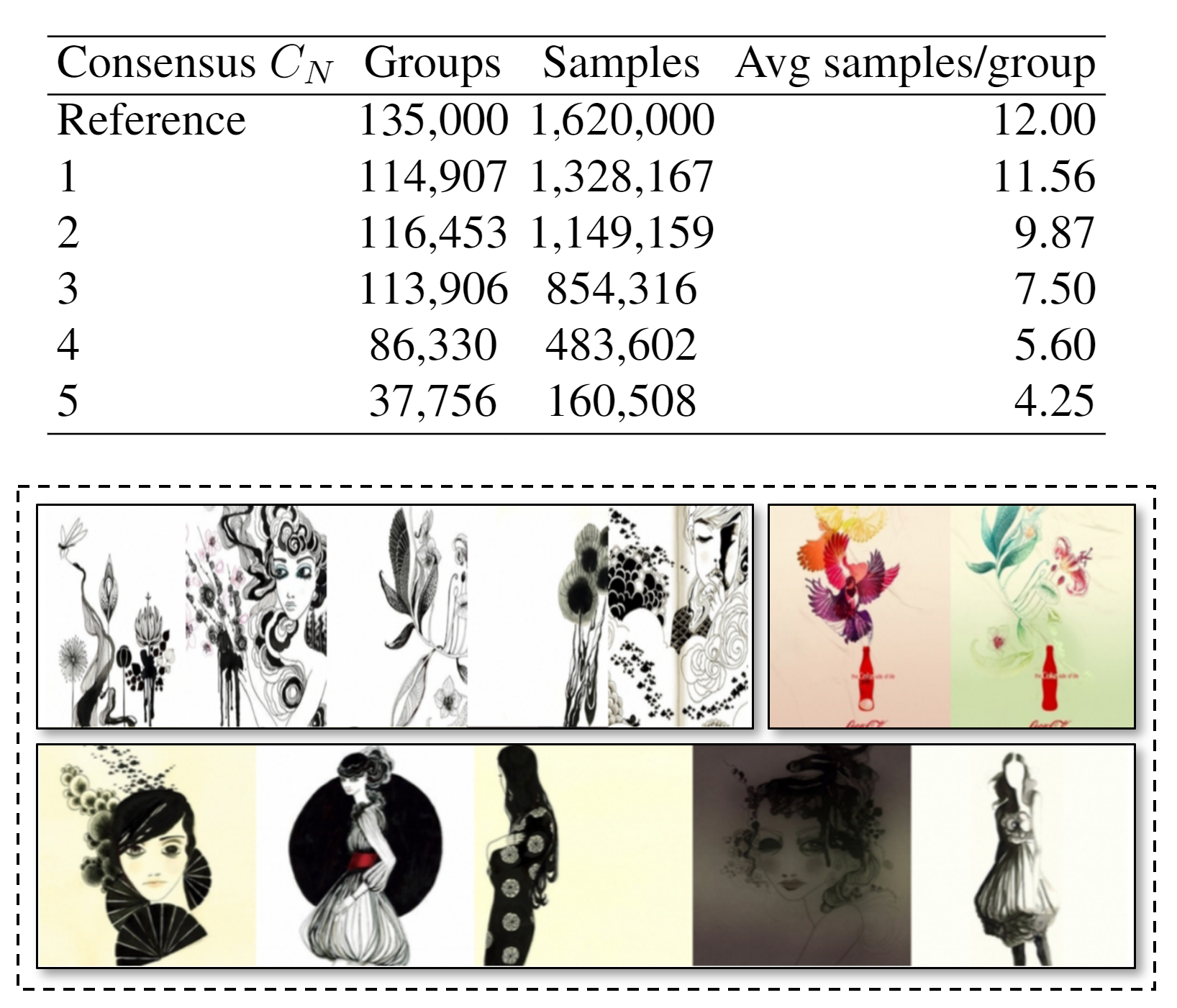}
    \caption{Top: BAM-FG-$C_N$ dataset statistics, cleaned at each worker consensus level $C_N=[1,5]$.  Bottom: Three sub-groupings within a single raw project group, identified at consensus level $C_N=3$
  }
    \label{fig:bamfg}
    \squeezeup
    \squeezeup
\end{figure}

We propose BAM-FG, a novel dataset of 2.62 million digital artworks in 310K project groups sampled from Behance.net -- a creative portfolio website.  We assume that image co-occurrence within these groups implies a weak cue for style similarity, and it is this signal we harness to train our model. We split BAM-FG into two distinct partitions:

{\bf BAM-FG-Raw.} The raw, 'noisy' data where images are grouped as on Behance; we refer to this dataset as BAM-FG-Raw comprising 1M images in 175K projects.  We train our model with this data.

{\bf BAM-FG-$C_N$}. Data that has been `cleaned' via a large-scale crowd-sourcing exercise so that image groups are known (rather than assumed) to be style consistent.  This dataset initially comprises 1.62M images and 135K groups.  We derive image groupings from this data to varying levels of confidence (participant consensus), written as $C_N=1,...,5$. Fig.  \ref{fig:bamfg} (top) describes the number of images (samples) in BAM-FG at each consensus level, where $C_N=5$ returns the highest data quality but lowest data volume. This data is used to evaluate ALADIN and fine-tune its performance (train/test split details are in subsec.~\ref{sec:evaldset}).

\subsubsection{Cleaning Style Groups (BAM-FG-$C_N$)}

We constructed BAM-FG-$C_N$ by manually curating the 135K project groups into sub-groups of coherent style. Annotation was crowd-sourced via Amazon Mechanical Turk (AMT) using 1073 workers. Workers were presented with images from a project and invited to tag any number of images (including zero) sharing the same visual style (creating a group).  If there were multiple styles present, they picked the largest group. We sent each annotation task to 5 workers. The consensus was determined using a graph-based vote pooling method in which edges coded by an affinity matrix $A_{i,j}$ reflecting the number of times both images $\{i,j\}$ were simultaneously selected within an annotated cluster/moodboard.  Thresholding $A_{i,j}$ at a given consensus level $C_N=[1,5]$ partitions the group into sub-groups. Fig.  \ref{fig:bamfg} (bottom) shows an example of three sub-groups formed at $C_N=3$ from a single raw group.  The supplementary material contains a task example and a visualization of the graph-vote algorithm. 

\begin{table*}[t!] 


      \centering
      \small
      \begin{tabular}{l|ccc|ccc||cccc|}
        \hline
        & \multicolumn{6}{c||}{Coarse grained} & \multicolumn{3}{c}{Fine grained}\\
            & \multicolumn{3}{c|}{BAM}  & \multicolumn{3}{c||}{BAM-X}  & \multicolumn{3}{c}{BAM-FG} \\
      Model        & Top-1  & Top-3  & mAP  & Top-1  & Top-3  & mAP  &  IR T-1  & IR T-5  & IR T-10  \\
        \hline
        Karayev \cite{Karayev2014}  & 29.95 & 36.44   & 0.34   & 21.62   & 23.06   & 0.18 & 3.28   & 5.43   & 6.59 \\
        NST \cite{gatys}   & 36.56   & 39.43   & 0.38 & 31.01   & 32.03   & 0.28 & 6.31   & 7.82   & 9.86 \\
        NST-CGM \cite{gmdotcos}   & 34.74   & 42.96   & 0.36 & 32.85   & 36.84   & 0.31 & 3.54   & 8.32  & 10.89 \\
        DML-BAM \cite{Collomosse2017} & 93.16   & 99.32   & 0.61 & 67.12 & 88.69 & 0.50 &  3.57 &  6.87   & 9.95 \\
        DML-BAMX \cite{Collomosse2017} &  98.16   & {\bf 99.95}   & 0.49 & 79.97 & 92.68 & {\bf 0.69} &  3.04 &6.18 & 8.70 \\
        ResNet50 \cite{he2016deep} & 73.59 & 96.01 & 0.162 & 51.05 & 78.01 & 0.163  &  1.97 & 5.18 & 8.41 \\
        MUNIT-Unsup~\cite{munit}& 33.79 & 45.05 & 0.44 & 45.76  & 74.41  & 0.22 & 4.12 & 6.92 & 9.45 \\
        MUNIT-Triplet~\cite{munit} & 43.29 & 46.47 & 0.589 & 66.28  & 88.66  & 0.271 & 17.68 & 25.89 & 31.67 \\
        MUNIT-Listwise ~\cite{munit} & 42.63 & 46.45 & 0.404 & 70.98  & 91.85  & 0.226   & 18.04 & 24.91 & 29.62 \\
        MUNIT-Contrastive~\cite{munit}& \textcolor{new}{53.80} & \textcolor{new}{86.36} & \textcolor{new}{0.284} & \textcolor{new}{27.87} & \textcolor{new}{58.49} & \textcolor{new}{0.137} & \textcolor{new}{6.43} & \textcolor{new}{10.02} & \textcolor{new}{12.96}   \\
        \midrule
        ALADIN-Unsup. & \textcolor{new}{93.86} & \textcolor{new}{99.79} & \textcolor{new}{0.394} & \textcolor{new}{83.13} & \textcolor{new}{96.66} & \textcolor{new}{0.308} & \textcolor{new}{11.48} & \textcolor{new}{16.38} & \textcolor{new}{19.84} \\

        \midrule

        ALADIN-Triplet & \textcolor{new}{96.59} & \textcolor{new}{99.90} & \textcolor{new}{0.638} & \textcolor{new}{78.60} & \textcolor{new}{95.28} & \textcolor{new}{0.336}   & \textcolor{new}{26.70} & \textcolor{new}{35.08} & \textcolor{new}{40.47} \\

        ALADIN-Listwise & \textcolor{new}{95.59} & \textcolor{new}{99.90} & \textcolor{new}{0.550} & \textcolor{new}{76.16} & \textcolor{new}{94.12} & \textcolor{new}{0.316}  & \textcolor{new}{37.90}  & \textcolor{new}{47.88} & \textcolor{new}{53.83}  \\

        ALADIN-Contrastive & \textcolor{new}{\textbf{99.48}} & \textcolor{new}{\textbf{99.95}} & \textcolor{new}{\textbf{0.737}} & \textcolor{new}{\textbf{85.28}} & \textcolor{new}{\textbf{98.07}} & 0.479 & \textcolor{new}{\textbf{56.89}} & \textcolor{new}{\textbf{66.80}} & \textcolor{new}{\textbf{71.94}}\\

        \bottomrule

      \end{tabular}

  \caption{Coarse (BAM/BAM-X) and Fine grained (BAM-FG) style discrimination for the ALADIN model compared to baselines on coarse and fine-grained retrieval (BAM-FG). The larger ALADIN-L model is used.
  \label{tab:baselines}
  }
\end{table*}
\section{Experiments and Discussion}
We evaluate the performance of ALADIN for both coarse and fine-grained retrieval tasks. We use a learning rate of $10^{-4}$ with decay 0.9 and the ADAM optimizer on a single NVidia Titan-X 12GB GPU for all experiments. We trained all models to convergence with early stopping.

\subsection{Datasets and Partitions}
\label{sec:evaldset}

\noindent{\bf BAM-FG/-Raw/-$C_N$.} We train all our models from scratch using 1M images with noisy grouping (BAM-FG-Raw). We split the 1.62M clean images in BAM-FG-$C_N$ into training (115K  groups) and test (20K groups) partitions. BAM-FG-$C_N$ is crowd-annotated to form sub-groups as described in Section~\ref{sec:dataset}.  The data is thresholded at different consensus levels $C_N=[1,5]$
(groupings referred to as BAM-FG-$\mathrm{C}_{1..5}$).
Fine-grained retrieval is evaluated over the 78K images test of $BAM-FG-C_3$; we use $C_3$ as the majority consensus level with the highest data volume.

\noindent{\bf BAM/-X.} Behance Artistic Media (BAM) is a public dataset of 2M contemporary artworks gathered from Behance.net and annotated using a large-scale active learning pipeline \cite{bam}. We sample a 70k subset of this dataset, the annotations for which include seven coarse style labels for non-photographic media (3D renderings, comics, pencil/graphite sketches, pen ink, oil paintings, vector art, watercolor).   To increase the coarse-scale category count on BAM, we add a further 70K images gathered from web photo collections focusing on photographic styles (luxury, neon, minimalist photography, metallic, abstract shots, geometric forms, pastel shades), bringing the total to 14 coarse style classes each of 10K images. This Extended BAM (BAM-X) dataset of 140K images is used for coarse-grain evaluation. None of these are included in BAM-FG.

\subsection{Baseline Methods and Losses}
\label{sec:coarsefinegrained}
\textbf{1. Methods.} We compare against several existing style representations.  \textbf{Karayev} \etal  sample activations from a late FC layer of CaffeNet \cite{Karayev2014}.  Neural Style Transfer \cite{gatys} \textbf{NST}  features  using VGG-19 \cite{vgg} pre-trained on ImageNet \cite{ Lin2016} are compared with a variant taking cosine of Gram matrices \textbf{NST-CGM} across layers \cite{gmdotcos}.  A baseline discriminative network \emph{ResNet50} \cite{he2016deep}, extracting the style representation (embedding) from the penultimate FC layer (2048-D) of a ResNet50 network.  We compare also against the coarse-grain style embedding of Collomosse \etal  \cite{Collomosse2017} applying Deep Metric Learning (DML) via triplet loss.  Their model \textbf{DML-BAM}  was trained on BAM \cite{bam}, and we retrain it also on BAM-X \textbf{DML-BAMX}. We compare against the style codes available from MUNIT  \cite{munit} which  explicitly trains a disentangled representation for stylization.  We evaluate MUNIT style codes for search, trained unsupervised (\textbf{MUNIT-Unsup} \cite{munit}) and trained via our contrastive scheme, per ALADIN (\textbf{MUNIT-Contrastive}).

\begin{table}[t!]  
  \centering
  \begin{adjustbox}{width=0.5\textwidth}

      \centering
      \begin{tabular}{lllcr|rr||rrr}
        \toprule
         & & & & & \multicolumn{2}{c||}{Coarse-grained} & \multicolumn{3}{c}{Fine-grained} \\
        Loss         & Backbone & Aug. & MLP  & \thead{Sup.} &  \thead{BAM \\ mAP} & \thead{BAMX \\ mAP} & IR-1 & IR-5  & IR-10 \\
        \midrule
        Triplet & MUNIT~\cite{munit} & & &  \checkmark& 0.589 & 0.271  & 17.68 & 25.89 & 31.67 \\
        Triplet & MUNIT~\cite{munit} & & & HN & 0.483 & 0.255  & 12.96 & 19.48 & 24.58 \\
        Triplet & MUNIT~\cite{munit} & & \checkmark & \checkmark & 0.482 & 0.252  & 7.39 & 12.70 & 17.16 \\
        Triplet & MUNIT~\cite{munit} & \checkmark & &  & 0.501 & 0.259  & 7.79 & 12.81 & 16.95 \\
        Listwise & MUNIT~\cite{munit} & & &  \checkmark & 0.447 & 0.244 & 20.15 & 27.40 & 32.27 \\
        Listwise & MUNIT~\cite{munit} & & \checkmark & \checkmark & 0.404 & 0.226 & 18.04 & 24.91 & 29.62 \\
        Listwise & MUNIT~\cite{munit} & \checkmark & & & 0.410 & 0.211 & 16.03 & 22.61 & 27.23 \\
        Contrastive& Resnet50~\cite{he2016deep}& & \checkmark & \checkmark & 0.231 & 0.143 & 40.14 & 50.90 & 57.19  \\
        Contrastive& Resnet50~\cite{he2016deep}  & \checkmark & \checkmark & & 0.209 & 0.104 & 28.51 & 36.72 & 41.96  \\
        \midrule
        Triplet & ALADIN-S  & & & \checkmark & 0.558 & 0.297 & 25.60 & 33.71 & 38.93\\
        Triplet & ALADIN-L  & & & \checkmark & 0.638 & 0.336 & 26.70 & 35.08 & 40.47\\ 
        Triplet & ALADIN-L  & & & HN & \textcolor{new}{0.609} & \textcolor{new}{0.312} & \textcolor{new}{23.75} & \textcolor{new}{31.87} & \textcolor{new}{37.10} \\ 
        Triplet & ALADIN-L  & & \checkmark & \checkmark & \textcolor{new}{0.614} & \textcolor{new}{0.335} & \textcolor{new}{22.12} &  \textcolor{new}{31.55} & \textcolor{new}{37.60} \\ 
        Triplet & ALADIN-L  & \checkmark & & & \textcolor{new}{0.417} & \textcolor{new}{0.214} & \textcolor{new}{21.07} & \textcolor{new}{28.20} & \textcolor{new}{32.81} \\

        Listwise & ALADIN-L & & & \checkmark & \textcolor{new}{0.463} & \textcolor{new}{0.247} & \textcolor{new}{25.14} & \textcolor{new}{33.44} & \textcolor{new}{38.80} \\ 
        Listwise & ALADIN-S & & \checkmark & \checkmark & 0.523 & 0.284& 30.03 & 38.42 & 43.67 \\
        Listwise & ALADIN-L & & \checkmark & \checkmark & 0.550 & 0.316 & 37.90 & 47.88 & 53.83 \\

        Listwise & ALADIN-L  & \checkmark & & & \textcolor{new}{0.453} & \textcolor{new}{0.244} & \textcolor{new}{21.22} & \textcolor{new}{27.98} & \textcolor{new}{32.36} \\ 
         Contrastive        & ALADIN-S  & & \checkmark & \checkmark & 0.565 & 0.308 & 36.42 & 44.49 & 49.32 \\

        Contrastive        & ALADIN-L  & & \checkmark & \checkmark & \textcolor{new}{\textbf{0.737}} & \textbf{0.479} & \textcolor{new}{\textbf{56.89}} & \textcolor{new}{\textbf{66.80}} & \textcolor{new}{\textbf{71.94}} \\
        Contrastive        & ALADIN-L  & \checkmark & \checkmark &  & \textcolor{new}{0.443} & \textcolor{new}{0.222} & \textcolor{new}{30.47} &  \textcolor{new}{38.43} & \textcolor{new}{43.39} \\
        \bottomrule
      \end{tabular}
  \end{adjustbox}
  \caption{Coarse and fine-grained style discrimination of different architectures and training losses, exploiting data augmentation (Aug.) and projection networks (MLP) under self or weak supervision (Sup.). HN indicates hard negative mining. Trained over 1M images in BAM-FG-Raw.
  }
  \squeezeup
  \squeezeup
  \label{tab:training_types}
\end{table}

\textbf{2. Alternative losses.} We also evaluate alternative training strategies for ALADIN, and the baselines.  Both {\bf Triplet} loss and {\bf Listwise} loss \cite{listwise} are explored as smaller batch alternatives to contrastive training \eg \textbf{MUNIT-Triplet}, and \textbf{MUNIT-Listwise}.  For triplet training we train using random negative mining, sampling $G^-$ from random projects, and also hard-negative mining (HN) sampling images from semantically similar projects.  For HN, $n \in G^-$ are samples that satisfy a semantic threshold $|S(n)< S(a) + \sum{p \in G^+}S(p)| < T$ set empirically to encourage disentanglement similar to deep metric learning (DML) over BAM \cite{Collomosse2017}.  We use a pre-trained auxillary ResNet/ImageNet embedding for $S(.)$.  The listwise loss \cite{listwise} is a differentiable approximation to mean Average Precision (mAP), recently applied to deep metric learning for search embeddings.  Using list-wise loss several style groups are chosen at random, and all samples are added as query (anchor) images to the batch $B=\{a_1,...,a_{32}\} \subset T$. For each $a$ we define $G^+$ as images co-present in the project, and $G^- = T \setminus G^+$. The listwise loss rewards higher ranking of $G^+$ versus $G^-$.

\begin{figure}[t!]
    \centering
    \includegraphics[width=0.8\linewidth,height=9cm]{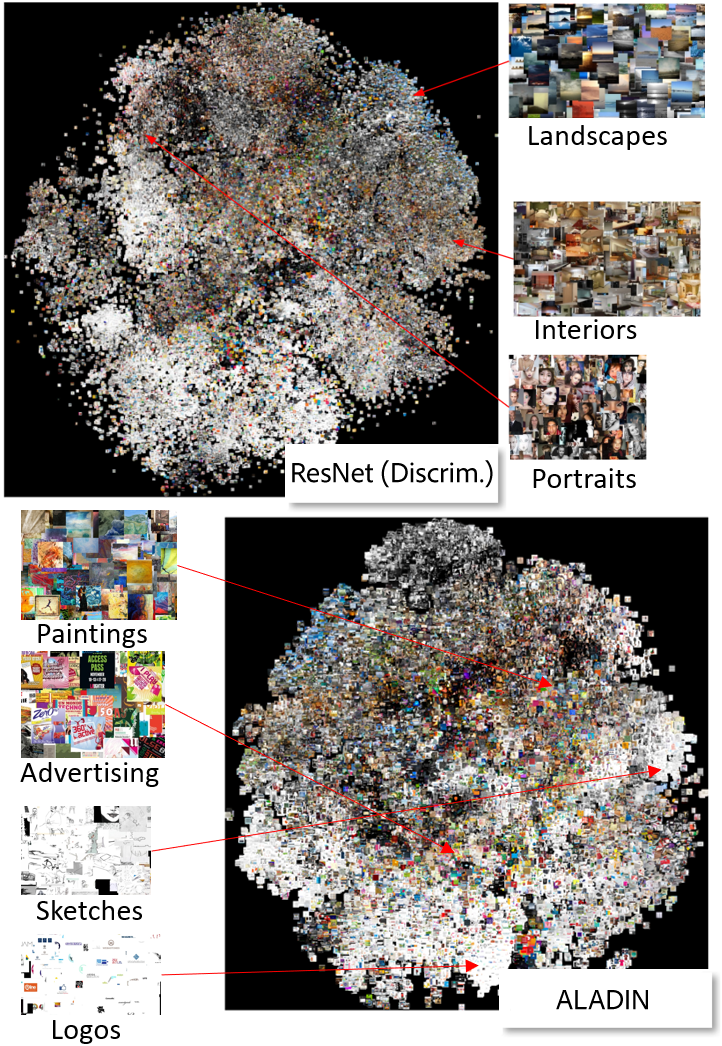}
    \caption{t-SNE visualization of BAM-FG test set within the ALADIN and discriminative embeddings; qualitatively fewer semantic clusters  form in the ALADIN vs. Discriminative embedding.}
    \label{fig:tsnebig}
    \squeezeup
\end{figure}

\begin{figure}[t!]
    \centering
    \includegraphics[width=0.9\linewidth,height=8cm]{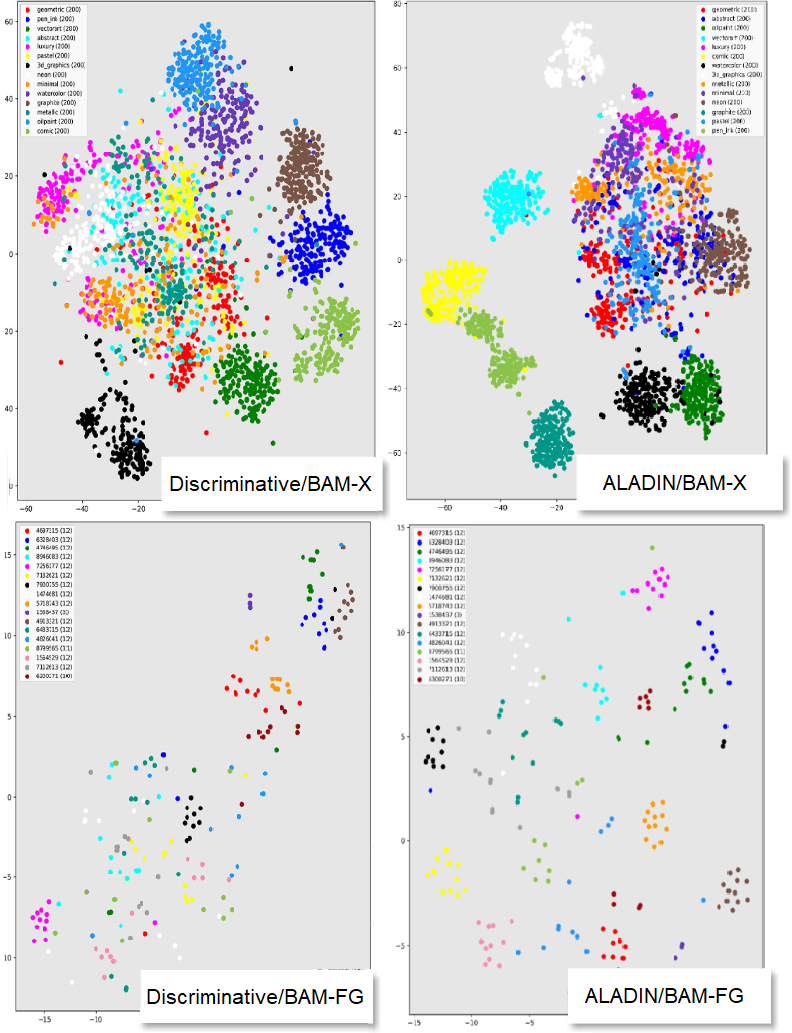}
    \caption{t-SNE visualizations in style embeddings (Discriminative and ALADIN-L). Top: BAM-X coarse grain, color shows 14 test classes. Bottom: BAM-FG, color shows 20 test projects.}
    \label{fig:tsne-cg-fg}
    \squeezeup
    \squeezeup
\end{figure}

\textbf{3. Alternative supervisions.}  We baseline against self-supervised training of ALADIN using no project group supervision \ie  $L_{\mathrm{total}}=L_{\mathrm{rec}}$ with minibatches selected from groups randomly.   Whilst self-supervised, such loss does not encourage metric learning.  Therefore we also baseline our dual loss eq. \ref{eq:dual}, where $L_{\mathrm{con}}$ uses vanilla contrastive learning (\ie self-supervised SimCLR \cite{simclr}), applying data augmentation to generate $G^+$ from $a$ using random cropping, horizontal flipping, and color jittering, and $G^-=B \setminus \{a,G^+\}$. Finally, we demonstrate the benefit of contrastive learning for weakly supervised learning, by baselining performance against {\bf softmax} loss for an $n-$way classification using project group membership as labelled data.   Thus we use our weak supervision signal as explicit labels for strong supervision, a 175K-way classification (one label per group).

\subsection{Evaluating Coarse and Fine grained Retrieval}

We evaluate retrieval performance of ALADIN, computing  \textbf{IR Top-$k$}; an instance retrieval (IR) metric that expresses the percentage of queries for which a relevant result (same project) occurs within the top $k$ ranked results, for ranks $k=\{1,5,10\}$. Tbl. \ref{tab:baselines} reports the performance of several baselines at this task; whilst coarse-grain performance is satisfactory on the BAM, and BAM-X datasets, the fine-grained style discrimination (BAM-FG) and generalization to new styles is poor for the baseline approaches, which supports our motivation to develop a new model (ALADIN) that is effective for fine grained style search.  The style code learned in {\bf MUNIT-UnSup}~\cite{munit} model is known to be suitable for image stylization, and we explore it here for style search.  We observe that for coarse-grain style discrimination, MUNIT performs similar (BAM IR-1: 33.79) to the lower-performing prior work that does not explicitly train for style   (BAM IR-1: 34.74) \cite{gmdotcos}. However {\bf MUNIT-UnSup} performs poorly as a search embedding for fine-grained style, in line with all prior models.  Our proposed approach, {\bf ALADIN-Contrastive} (using the ALADIN-L backbone), sets the new state of the art in fine grained retrieval (BAM-FG IR-1: 56.89).


Tbl~\ref{tab:baselines} also presents the results for coarse and fine grain for our proposed ALADIN network trained unsupervised (via reconstruction loss only; {\bf ALADIN-UnSup}) on BAM-FG\_Raw. The fine grained IR performance (BAM-FG IR-1: 26.70) of ALADIN-L exceeds all prior baseline architectures even without supervision.

\subsection{Evaluating weak supervision}

Next, we evaluate different ways to weakly supervise ALADIN training on BAM-FG-Raw noisy image groupings, comparing triplet, listwise and contrastive losses (Tbl. \ref{tab:training_types}). The ALADIN model (the ALADIN-L model structure) shows gains at both coarse-grain discrimination (+ 10-12\% mAP), and as shown in Tbl~\ref{tab:baselines}, fine-grained style discrimination (+ 10-15\% IR Top-1 ). Our proposed dual loss (reconstruction and contrastive learning) outperform others with listwise loss gives superior performance to triplet --- reflecting observations that even human annotators find it harder to assess style coherence in paired images versus groups of images.  A discriminative Resnet50 backbone was also trained using a softmax loss, with a $\sim$175k-way classifier for every project group in the 1M BAM-FG-Raw dataset. However, it did not perform well for fine-grained discrimination.  ALADIN achieves a state of the art performance on fine-grained style discrimination, but interestingly outperforms existing models for coarse grain style. We achieve this without using any of the coarse training labels, demonstrating our model's generalization capabilities.

\subsection{Ablation Studies}
\label{sec:eval}

\begin{table}
  \centering
  \begin{adjustbox}{width=0.45\textwidth}

      \centering
      \begin{tabular}{ll|ccc|ccc}
        \toprule
                     & Data & \multicolumn{3}{c|}{ResNet50 \cite{he2016deep}} & \multicolumn{3}{c}{ALADIN-L} \\
        Data   & size & IR-1 & IR-5 & IR-10 & IR-1 & IR-5 & IR-10 \\
        \midrule
        BAM-FG-Raw   & 250K & 20.72 & 28.72 & 34.30 & \textcolor{new}{40.08} & \textcolor{new}{49.62} & \textcolor{new}{55.27} \\ 
        BAM-FG-Raw   & 500K & 33.80 & 43.74 & 49.71 & \textcolor{new}{44.26} & \textcolor{new}{54.20} & \textcolor{new}{59.75} \\ 
        BAM-FG-Raw   & 750K & 38.73 & 49.11 & 55.09 & \textcolor{new}{47.78} & \textcolor{new}{57.89} & \textcolor{new}{63.53} \\ 
        BAM-FG-Raw   & 1M & 40.14 & 50.90 & 57.19 & \textcolor{new}{56.89} & \textcolor{new}{66.80} & \textcolor{new}{71.94}  \\ 
        \midrule
        BAM-FG-$\mathrm{C}_3$ & 114K & 42.33 & 52.14 & 57.72 & 59.03 & 68.67 & 73.62 \\ 
        BAM-FG-$\mathrm{C}_4$ & 86K & 43.85 & 53.86 & 59.37 & 59.31 & 68.80 & 73.92 \\ 
        BAM-FG-$\mathrm{C}_5$ & 38K & 45.22 & 54.81 & 60.52 & \textbf{59.51} & \textbf{69.07}  & \textbf{74.05} \\
        \bottomrule
      \end{tabular}

  \end{adjustbox}
  \caption{The effect of data volume and consensus level on a fine-tuned model, training with contrastive loss.
  }
  \label{tab:data_volume_tab}


  \centering
  \small

      \centering
      \begin{tabular}{ll|rrr}
        \toprule
        Data & Model &  IR-1 & IR-5 & IR-10 \\
        \hline
        BAM & DML \cite{Collomosse2017} & 3.57 & 6.87 & 9.95 \\
        \midrule
        BAM-FG-C$_5$ & Ours (ALADIN-L) & \textcolor{new}{58.98} & \textcolor{new}{68.79}  & \textcolor{new}{73.73} \\
        BAM-FG-C$_5$ & Ours (ResNet50 \cite{he2016deep})& 45.22 & 54.81 & 60.52 \\
        \midrule
        BAM-FG-Raw   & Ours (Fused) & \textcolor{new}{60.76} & \textcolor{new}{70.26} & \textcolor{new}{75.12} \\
        BAM-FG-C$_5$ & Ours (Fused) & \textcolor{new}{\textbf{62.18}} & \textcolor{new}{\textbf{71.50}} & \textcolor{new}{\textbf{76.33}} \\
        \bottomrule
      \end{tabular}
  \caption{Performance of fused embeddings Discriminative and ALADIN for noisy (BAM-FG-Raw) and cleaned (BAM-FG).}
  \label{tab:fused}
  \squeezeup
\end{table}
\begin{figure*}[t!]
     \centering
     \includegraphics[width=0.85\linewidth,height=6.3cm]{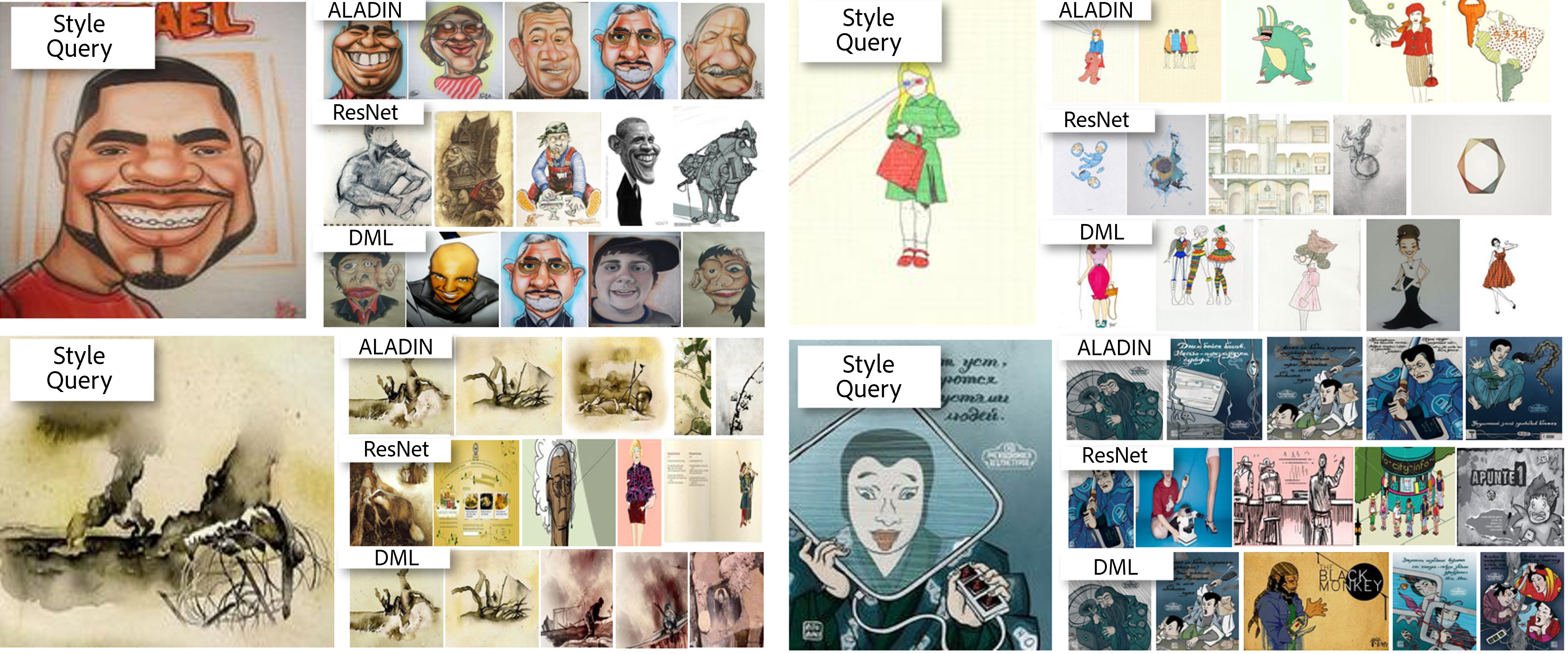}
     \caption{Representative retrieval results for 4 queries;
    Coarse-grain style embedding baseline  (DML) \cite{Collomosse2017} versus our proposed fine-grained embedding (ALADIN architecture under weak supervision) and for comparison, a similarly trained (discr)iminative network; ResNet.
     }
     \label{fig:gallery}
 \end{figure*}
Tbl~\ref{tab:training_types} reports an ablation study for both coarse and fine grained style discrimination, exploring contrastive learning, data augmentation and use of a projection network (applying the loss to $h(f(.))$ versus $f(.)$), the latter yields performance gain for both models. The pair-wise (triplet) losses consistently underperform the list-wise loss. Performing hard negative mining to encourage disentanglement between semantics and style is ineffective, perhaps due to the disentanglement in the ALADIN architecture. ALADIN-L  (56.89\%)  outperforms baselines for fine-grained discrimination, including a recent self-supervised baseline (SimCLR \cite{simclr}). In contrast, a discriminative model~\cite{he2016deep}, ResNet50, similarly trained, achieves 40.14\%. 

Tbl. \ref{tab:data_volume_tab} explores the impact of data volume on overall performance for reductions of 25-75\% of BAM-FG-Raw data, for the best performing models ALADIN-L (IR-1: 56.89\%) and Discriminative (IR-1: 40.14\%). We also explored fine-tuning these two models with the cleaned BAM-FG training data at consensus levels $C_N=[3,4,5]$. Despite lower training volumes at higher consensus levels, we show the most significant gains are at these levels (up to 5\% at $C_5$).

\subsection{Visualizing Disentanglement}

We explore the complementary properties of ALADIN-Contrastive and a similarly trained discriminative network (ResNet50-Contrastive) that does not explicitly disentangle style from content in its architecture.  Fig.~\ref{fig:gallery}, shows search results for both embeddings, offering a more significant correlation between content and style under the Discriminative model, which does not explicitly disentangle the two within its architecture.

Visualizing a t-SNE distribution of all 78K images in BAM-FG-$\mathrm{C}_3$ (Fig. \ref{fig:tsnebig}) shows similar groupings. It reveals ALADIN correlates more strongly on lower level features such as color.
Smaller-scale visualization of Fig.~\ref{fig:tsne-cg-fg} shows good discrimination between a random subset of 20 fine-grained groups and on coarse-grain styles despite no direct supervision on these classes; consistent with the quantitative performance in Tbl.~\ref{tab:baselines}.


\noindent {\bf Fused.} We explored simply concatenating both embeddings given their information appeared complementary. Interestingly, the incorporation of semantic cues gave a small performance boost (around $4\%$) to ALADIN. Further small gains can be made by further fine-tuning ALADIN on cleaned annotated data from BAM-FG-$C_N$.   Tbl. \ref{tab:fused} reports our
best performing fine-grained search embeddings are the fused models on noisy data (60.76\%) and cleaned data at the highest consensus threshold ($C_5$ at 62.18\%).

\begin{table*}

 \centering
\begin{adjustbox}{width=0.70\linewidth}
 \begin{tabular}{l|ccc|ccc|ccc|ccc}
    \toprule
        Consensus & \multicolumn{3}{c|}{ResNet50 \cite{he2016deep}}  & \multicolumn{3}{c|}{Ours (ALADIN)} & \multicolumn{3}{c|}{Ours (Fused)}  & \multicolumn{3}{c}{DML-BAMX \cite{Collomosse2017}} \\
    $C_{N}$  & P@1 & P@5 & P@10 & P@1 & P@5 & P@10 & P@1 & P@5 & P@10 & P@1 & P@5 & P@10 \\
    \midrule
    3  & 0.959 & 0.914 & 0.895 & 0.965 & 0.918 & 0.893 & \textbf{0.974} & \textbf{0.937} & \textbf{0.918} & 0.780 & 0.607 & 0.449 \\
    4  & 0.840 & 0.681 & 0.624 & 0.832 & 0.653 & 0.582 & \textbf{0.867} & \textbf{0.727} & \textbf{0.666} & 0.633 & 0.397 & 0.277 \\
    5  & 0.562 & 0.324 & 0.256 & 0.612 & 0.307 & 0.221 & \textbf{0.642} & \textbf{0.363} & \textbf{0.277} & 0.467 & 0.196 & 0.120 \\
    \midrule
    3 (c) & 0.822 & 0.686 & 0.519 & 0.834 & 0.831 & 0.826 & \textbf{0.887} & \textbf{0.890} & \textbf{0.875} & 0.681 & 0.571 & 0.428 \\
    4 (c) & 0.522 & 0.432 & 0.322 & 0.558 & 0.458 & 0.450 & \textbf{0.575} & \textbf{0.569} & \textbf{0.560} & 0.364 & 0.307 & 0.229 \\
    5 (c) & 0.153 & 0.117 & 0.096 & 0.115 & 0.106 & 0.101 & \textbf{0.173} & \textbf{0.166} & \textbf{0.159} & 0.079 & 0.065 & 0.049 \\
   \bottomrule
  \end{tabular}
  \end{adjustbox}
  \caption{\textcolor{new}{Image retrieval accuracy via Amazon Mechanical Turk (AMT) study; Precision (P@k) at rank k=[1,5,10] over 1000 queries on BAM-FG. (c) indicates "held out" evaluation i.e. removing results in the same style group as the query. These show that when images from the same group are removed, ALADIN (and fused) maintains style similar results for longer. Results at 3 majority consensus levels.}}
  \label{tab:img_ret_usersAG}

\end{table*}

\subsection{User Evaluation of Style Similarity}
\label{sec:user_study}

We ran crowd-sourced user studies on image retrieval results to compare ALADIN with the fine-grained ResNet-Contrastive model trained on the same data and a coarse-grain baseline (DML \cite{Collomosse2017}). 1000 queries were each presented independently to 5 workers. Each worker was asked to flag images within the 25 results that were style inconsistent with the query. Consensus thresholding (Sec.~\ref{sec:dataset}) was applied and Precision at rank $k$ recorded for majority consensus levels in Tab. \ref{tab:img_ret_usersAG}.
The fused embeddings consistently achieve the highest accuracy, both where we explore the complete results and when we consider only results where the query style group is held out from the search corpus.

\begin{figure}[t!]
    \squeezeup
    \centering
    \includegraphics[width=1.0\linewidth,height=5cm]{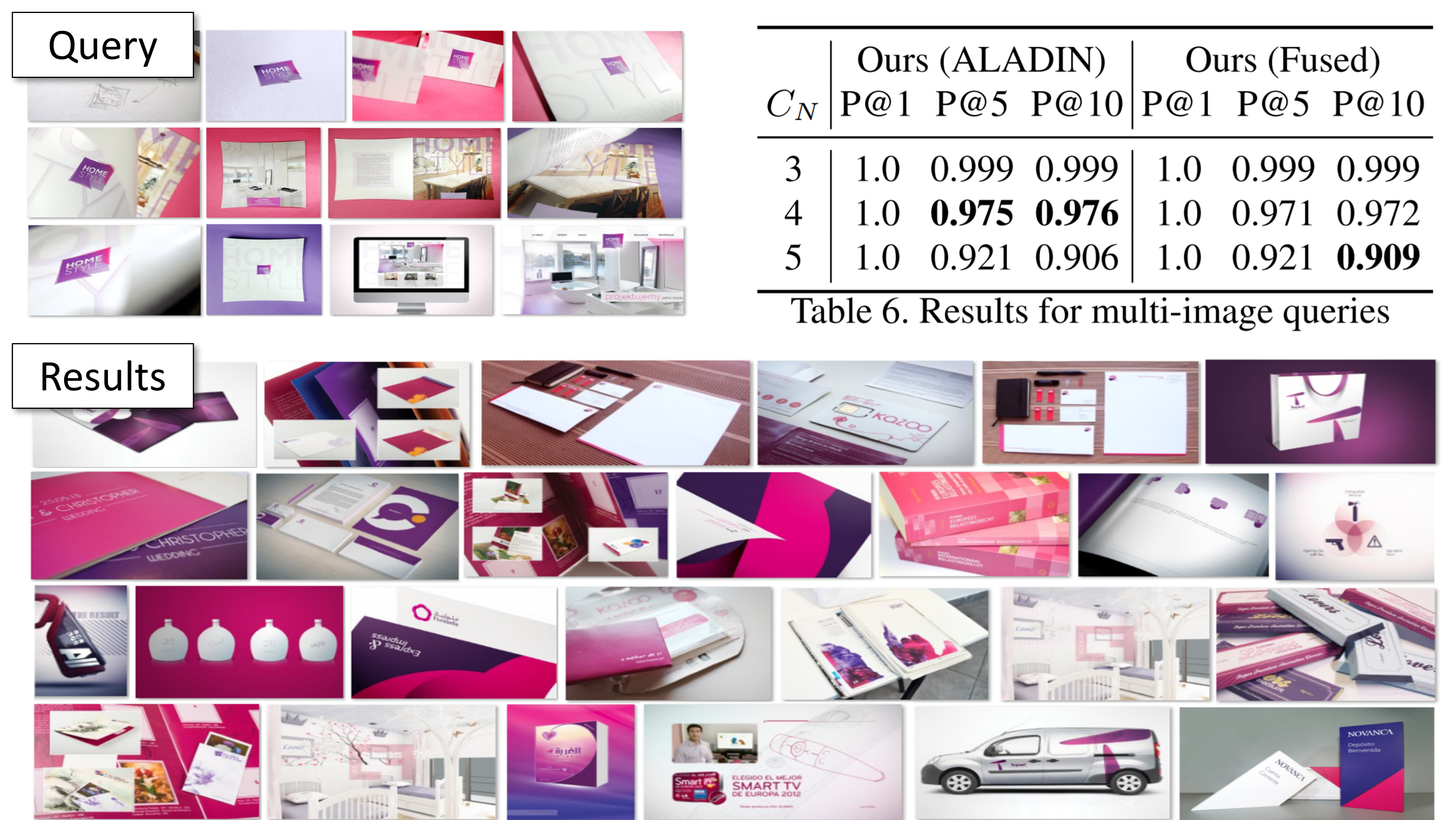}
    \caption{Multi-image query results; a query moodboard (top left) and the top 25 ranked result images (bottom). Quantiative results from MTurk user study at majority consensus levels (top right).}
    \label{fig:multiimage}
    \squeezeup
\end{figure}


We further explore image retrieval using the models when multiple images are used together to form a query. We use our annotated style groups from the test set as such queries, mean-pooling the encoder features of all images to create the query. We repeat our user study and ask groups of 5 workers to label 1000 retrieval results. The retrieval corpus does not contain images from the same project as the queries.
Fig.~\ref{fig:multiimage} shows multi-image queries, and a representative result with high relevance scores for both ALADIN and the fused variants, implying the mean well represents the query group and suitable for use as a query.


\vspace{-0.2cm}
\section{Conclusion}
\vspace{-0.1cm}

We proposed an encoder-decoder (E-D) architecture; `ALADIN', to learn a fine-grained representation for measuring fine-grained artistic style similarity. ALADIN is inspired by the E-D models used to drive content stylization \cite{munit} by explicitly disentangling content and style across network branches. We present and use a weakly supervised approach to learn this representation using contrastive training made practical using a logit accumulation strategy. We sample 310K user-generated content groupings across 2.62M images which we release as the first fine-grained artistic style dataset, BAM-FG. We exploit these group co-occurrences as an implicit weak proxy for fine-grained style similarity.  ALADIN sets the state of the art performance for fine-grained discrimination and improves further when trained with cleaned grouping data and when fused.  ALADIN also sets a new state of the art in coarse grain style discrimination, despite not being trained with explicit style labels.  Future work could explore the potential of ALADIN as a unified embedding for both stylization and search.




{\small
\bibliographystyle{ieee_fullname}
\bibliography{main}
}

\end{document}